\documentclass[10pt,twocolumn,letterpaper]{article}

\usepackage[pagenumbers]{cvpr} %

\usepackage[dvipsnames,table]{xcolor}

\usepackage{graphicx}
\usepackage{multirow}
\usepackage{makecell}
\usepackage{subcaption}

\newcommand{\myparagraph}[1]{\vspace{1pt}\noindent{\bf #1}}
\newcommand{\tablestyle}[2]{\setlength{\tabcolsep}{#1}\renewcommand{\arraystretch}{#2}\centering\small}
\newcommand{\thickhline}{\Xhline{3\arrayrulewidth}}
\newcommand{\gain}[1]{{\color{blue}\scriptsize\textbf{#1}}}
\newcommand{\unsim}{\mathord{\sim}}  %

\newcommand{\shortreffig}[1]{Fig.~\ref{fig:#1}}

\newcommand{\shortrefsec}[1]{\S\ref{sec:#1}}

\newcommand{\lblfig}[1]{\label{fig:#1}}

\usepackage{pgfplots}
\usepgfplotslibrary{fillbetween}
\usepackage{pgfkeys}
	\newenvironment{customlegend}[1][]{%
		\begingroup
		\csname pgfplots@init@cleared@structures\endcsname
		\pgfplotsset{#1}%
	}{%
		\csname pgfplots@createlegend\endcsname
		\endgroup
	}%
	\def\addlegendimage{\csname pgfplots@addlegendimage\endcsname}

\definecolor{black}{rgb}{0.0, 0.0, 0.0}
\definecolor{blue}{rgb}{0.11764705882352941, 0.5647058823529412, 1.0}
\definecolor{green}{rgb}{0.66,0.82,0.56}
\definecolor{darkgreen}{rgb}{0.545, 0.749, 0.608}
\definecolor{darkergreen}{rgb}{0.384, 0.631, 0.576}
\definecolor{orange}{rgb}{0.9568627450980393, 0.3176470588235294, 0.11764705882352941}
\definecolor{Gray}{gray}{0.9}

\def\vx{{\mathbf{x}}}
\def\vy{{\mathbf{y}}}

\definecolor{cvprblue}{rgb}{0.21,0.49,0.74}
\usepackage[pagebackref,breaklinks,colorlinks,citecolor=cvprblue]{hyperref}

\title{Training a Large Video Model on a Single Machine in a Day}

\author{Yue Zhao \quad Philipp Kr\"ahenb\"uhl\\
UT Austin\\
{\tt\small \{yzhao,philkr\}@cs.utexas.edu}
}

\begin{document}
\maketitle

\begin{abstract}
    Videos are big, complex to pre-process, and slow to train on.
    State-of-the-art large-scale video models are trained on clusters of $32$ or more GPUs for several days.
    As a consequence, academia largely ceded the training of large video models to industry.
    In this paper, we show how to still train a state-of-the-art video model on a single machine with eight consumer-grade GPUs in a day.
    We identify three bottlenecks, IO, CPU, and GPU computation, and optimize each.
    The result is a highly efficient video training pipeline.
    For comparable architectures, our pipeline achieves higher accuracies with $\frac{1}{8}$ of the computation compared to prior work.
    Code is available at \url{https://github.com/zhaoyue-zephyrus/AVION}.
\end{abstract}

\section{Introduction}

Video understanding has witnessed remarkable advances in the past decade.
Much of the progress on standard benchmarks~\cite{carreira2017i3d} is powered by higher-capacity models~\cite{carreira2017i3d,feichtenhofer2019slowfast,arnab2021vivit,yan2022multiview} trained on ever larger datasets~\cite{ghadiyaram2019ig65m,miech2019howto100m,stroud2020wts,yuan2021florence}.
The result is an ever increasing training cost, exacerbated by the recent shift from convolutional~\cite{carreira2017i3d,xie2018s3d,feichtenhofer2020x3d,kondratyuk2021movinets} to Transformer architectures~\cite{bertasius2021timesformer,arnab2021vivit,li2022mvitv2,liu2022videoswin,yan2022multiview}. 
For much of their evolution, video models followed the success of their image-based counterparts~\cite{he2016resnet,howard2017mobilenets,dosovitskiy2021vit,liu2021swin}.
However, working with videos offers a series of unique challenges: Videos are highly compressed, up to an order of magnitude more than images.
Video decoding consumes a sizeable fraction of the overall computation in state-of-the-art training pipelines~\cite{feichtenhofer2022maest}.
Finally, the decompressed soup of pixels grows not just quadratically with the input resolution, but also with temporal length.
This puts a strain on pre-processing pipelines and significantly increases the GPU memory that a video model uses.

In this paper, we examine the training pipeline of a modern video Transformer architecture~\cite{bertasius2021timesformer,lin2022egovlp} from three perspectives: model, video loading, and video pre-processing, which are GPU-, IO-, and CPU-bound respectively.
We find that there is plenty of room for improvement in all aspects.
Through careful designs we improve the training time by almost an order of magnitude.

From the model perspective, we start with the plain, non-hierarchical Vision Transformer and reduce the memory bottleneck from $ O(N^2) $ to $ O(N) $, where $ N $ is the length of the cubified video tokens.
We achieve this by adopting FlashAttention~\cite{dao2022flashattention} which decomposes the whole sequences into SRAM-friendly blocks and combines the block-wise results into the final output without explicitly storing the full attention weight matrix.
This results in a reduced per-video memory cost as well as an increased training throughput.
The reduced per-instance memory footprint enables training a video model with a significantly larger batch size on a single multi-GPU server.
This is particularly useful for training CLIP-style models~\cite{radford2021clip} for videos, which typically requires as many as $32\unsim64$ GPUs or TPUs~\cite{lin2022egovlp,miech2020milnce} to construct a batch of $\unsim1K$ video instances.

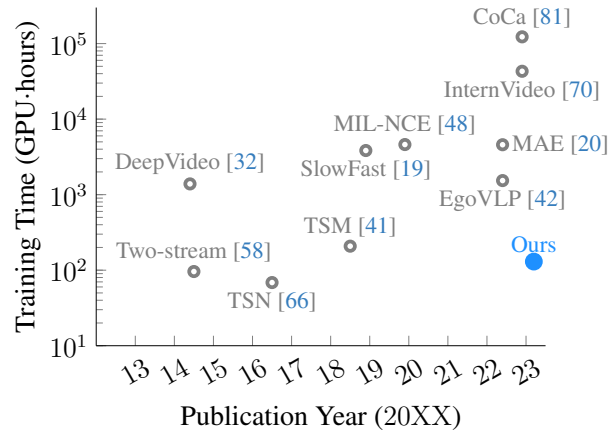
\begin{figure}[t]
    \centering
    {
        \resizebox{\linewidth}{!}{
        \begin{tikzpicture}
	\begin{axis} [
		axis x line*=bottom,
		axis y line*=left,
        ylabel={Training Time (GPU$\cdot$hours)},
        xlabel={Publication Year ($20\mathrm{XX}$)},
        legend pos=north east,
		ymin=10, ymax=3e5,
		xmin=12, xmax=23.5,
		xticklabel={\pgfmathparse{\tick}\pgfmathprintnumber{\pgfmathresult}},
		xtick={13, 14, 15, 16, 17, 18, 19, 20, 21, 22, 23},
		nodes near coords,
        nodes near coords style={
            anchor=south,
        },
        ylabel style={align=center, font=\large, at={(0.03, 0.5)}},
        xlabel style={align=center, font=\large, at={(0.5, -0.05)}},
        ymode=log, log origin=infty,
        xticklabel style={rotate=30},
		width=\linewidth,
		height=0.8\linewidth,
		tick label style={font=\large},
		]
		\addplot [only marks, mark=o,style={ultra thick},gray, point meta=explicit symbolic]  coordinates {
            (14.4, 1388) [DeepVideo~\cite{karpathy2014deepvideo}]
            (19.9, 4608) [MIL-NCE~\cite{miech2020milnce}]
            (22.9, 122880) [CoCa~\cite{yu2022coca}]
            (14.5, 96) [Two-stream~\cite{simonyan2014twostream}]
            (18.5, 208) [TSM~\cite{lin2019tsm}]
		};
		\addplot [only marks, mark=o,style={ultra thick},gray, point meta=explicit symbolic,nodes near coords style={
            anchor=north,
        },]  coordinates {
            (18.9, 3840) [SlowFast~\cite{feichtenhofer2019slowfast}]
            (16.5, 68.6) [TSN~\cite{wang2016tsn}]
            (22.4, 1536) [EgoVLP~\cite{lin2022egovlp}]
            (22.9, 43008) [InternVideo~\cite{wang2022internvideo}]
        };
		\addplot [only marks, mark=o,style={ultra thick},gray, point meta=explicit symbolic,nodes near coords style={
            anchor=west,
        },]  coordinates {
            (22.4, 4582) [MAE~\cite{feichtenhofer2022maest}]
        };
        \addplot [mark=*, mark options={scale=1.5, ultra thick}, blue, point meta=explicit symbolic] plot coordinates {
			(23.2, 130) [Ours]
		};
 \end{axis}
\end{tikzpicture}
        }
    }
    \vspace{-15pt}
    \caption{\small{Over the past decade training time of state-of-the-art video models increased by two orders of magnitude, despite drastic improvements in GPU hardware. State-of-the-art video models train on 6 GPU$\cdot$months to 14 GPU$\cdot$years of computation on cutting-edge hardware.
    We show how to train an equally performant large video model in under a day on a machine with eight workstation GPUs. (Metrics not normalized for GPU generations).}}
    \label{fig:teaser}
\end{figure}

The increased throughput, however, introduces additional challenges to video loading and pre-processing.
In our pipeline, we redesign the video loader around a series of trimmed fixed-length chunks of a long-form video.
Each chunk is still compressed using a modern video codec~\cite{richardson2011h264}.
The GPU hardware determines the length of each chunk.
A chunk-based representation reduces the IO bottleneck and increases the decoding speed.

We merge the commonly used \textrm{RandomResizedCrop} operation into the video decoding stage as a cropping filter.
This ensures the video decoder executes a minimal amount of decoding operations to retrieve the required data.
Furthermore, we move all other data augmentations to the GPU to make use of its massive parallelism.

We evaluate our pipeline on contrastive video-language pre-training on the Ego4D video-narrative pairs~\cite{grauman2022ego4d}.
Our pipeline trains a contrastive video-language model on 4M video-text pairs with a total batch size of $2K$ clips using \textbf{one} $ 8\times $ A5000 (24GB) GPU server in \textbf{18} hours.
The same model used to require $ 32\times $ 40GB A100 GPUs to run for 2 days~\cite{lin2022egovlp}.
Our pipeline leads to a $ 6.7\times $ reduction of memory consumption, $ 11.8\times $ reduction of GPU$\cdot$hours, and $ 15 \times $ reduction in hardware cost\footnote{We only compare GPUs' MSRP: One A5000 costs $ \unsim\$2,600 $ while one A100 costs $\unsim\$10,000$. Networking and distributed filesystem are likely to cost more for multi-node setup.}.
With an LLM-augmented set of video narratives~\cite{zhao2022learning}, our model is able to achieve state-of-the-art performance on Epic-Kitchens 100 Multi-Instance Retrieval in both zero-shot and fine-tuning evaluation protocols.
With a comparable model (ViT-Base \vs a TimeSformer-Base), our model is 2.0\% higher in terms of zero-shot average mAP and 1.3\% better after fine-tuning.

Our optimized pipeline as an application works beyond large video-language modeling.
We show additional results on training Video Masked Auto-Encoders (MAE) where data loading is a bottleneck, our techniques reduce the data-loading overhead by $ 3\times $ and overall training time by $ 35\% $.

\section{Related Work}

\myparagraph{Computationally Efficient Video Recognition.}
Video models that are inflated from image models through time are computationally expensive~\cite{carreira2017i3d}.
Architectural improvements include channel-wise separable convolution~\cite{tran2019csn}, temporal channel shuffling~\cite{lin2019tsm}, dilated temporal convolution~\cite{hussein2019timeception}, depth-parallel pipelining~\cite{carreira2018massively}, and progressive expansion of multiple network axes across space, time, width and depth~\cite{feichtenhofer2020x3d}.
Some works attempt to represent motion information using compressed videos~\cite{wu2018compressed,zhang2016real} or temporal difference~\cite{wang2021tdn} to avoid the expensive computation of optical flow in the two-stream network~\cite{simonyan2014twostream}.
Other works focus on reducing the spatial-temporal redundancy in videos by selecting informative frames~\cite{korbar2019scsampler,wu2019adaframe} or regions~\cite{pan2021iared,wang2021adafocus}, and quantization~\cite{meng2020arnet,sun2021dynamic}.
Training can be sped up by applying a multigrid schedule across variable spatial-temporal resolutions~\cite{wu2020multigrid} or curriculum strategy~\cite{bain2021frozen}.
Our contributions are complementary and focus on the IO and preprocessing bottlenecks on standard video training pipelines.
Our main architectural improvements are Transformers~\cite{vaswani2017attention} specific.

\myparagraph{Efficient Transformers.}
The dot-product attention in the original Transformer~\cite{vaswani2017attention} requires quadratic computation in the input sequence length.
This becomes prohibitive for long sequences, and many works focus on reducing this computation.
Representative approaches include low-rank approximation of the attention matrix~\cite{choromanski2021performers}, sparsity patterns~\cite{zaheer2020bigbird}, reversible transform~\cite{kitaev2020reformer}, query-based cross attention via memory~\cite{rae2020compressive} and recurrence~\cite{dai2019transformerxl}, and kernel decomposition~\cite{katharopoulos2020transformers}.
In video understanding, Transformers are tailored to modeling video sequences by (1) computing attention across separate dimensions~\cite{bertasius2021timesformer,patrick2021motionformer}, (2) building hierarchy through shifted local windows~\cite{liu2022videoswin} and multi-scale pooling attention~\cite{fan2021mvit}.
MemViT~\cite{wu2022memvit} models $30\times$ longer temporal support with a marginal extra cost via recurrence.
RevViT~\cite{mangalam2022revvit} reformulates the self-attention block with reversible transform~\cite{gomez2017revnet}.
TeSTra~\cite{zhao2022testra} adopts temporal-smoothing kernels to process streaming videos with both constant computation and memory per frame.
In contrast, we take a brute-force approach to the problem.
In video transformers, the quadratic memory consumption is a much larger problem, than the actual increase in computation.
We leverage recent advances in efficient implicit computation of the attention weights~\cite{jang2019mnnfast} implemented in a computationally efficient block-wise manner in FlashAttention~\cite{dao2022flashattention}.
FlashAttention eliminates the memory bottleneck and significantly increases the computation throughput of the attention operation.
The result is a ViT-base network that is as efficient as factorized representations, but uses a fraction of the memory.
Keeping the original ViT-base structure also allows us to make use of image-based pre-training either contrastive~\cite{radford2021clip,schuhmann2022laion5b} or self-supervised~\cite{caron2021dino,he2022mae}.

\myparagraph{Memory-Efficient Video Models.}
To fit longer video clips into GPUs, many approaches resort to extracting frame-level or short-clip features and building an additional model for temporal reasoning on top~\cite{miech2019howto100m,xu2021videoclip,zolfaghari2018eco}.
The performance of such models is heavily constrained by the representation capability from the frame-level model.
For end-to-end video models, efforts that aim to reduce memory footprint include sparse sampling frames~\cite{zhao2017ssn}, dropping out gradients~\cite{cheng2022stochastic}, and skipping intermediate blocks~\cite{wu2018blockdrop}.
However, most of them either focus on the inference stage or speed up training a particular family of models.
In contrast, our optimization on the IO- and CPU-bound operations should be applicable to all kinds of video models.

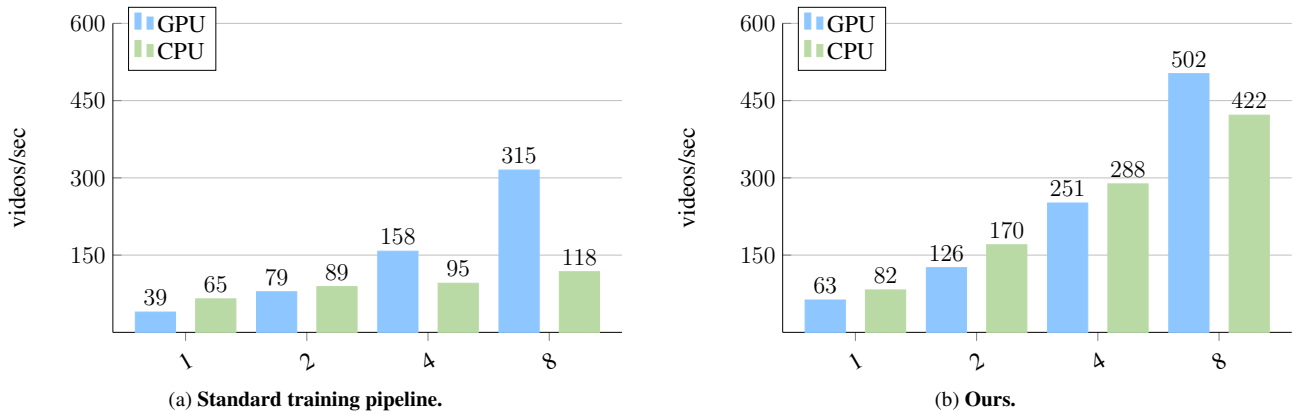
\begin{figure*}[!tb]
    \centering
    \subfloat[
	\textbf{Standard training pipeline.}
	\label{fig:bottleneck_baseline}
	]
    {
		\resizebox{0.48\linewidth}{!}{
		\begin{tikzpicture}
	\begin{axis}[
		ybar=10pt,
		bar width=20pt,
        enlarge x limits=0.2,
  		axis y line*=left,
		axis x line*=bottom,
		ylabel={videos/sec},
        symbolic x coords={1, 2, 4, 8},
		xtick=data,
        nodes near coords,
		nodes near coords align={vertical},
		every node near coord/.append style={font=\large, black, /pgf/number format/.cd,fixed zerofill,precision=0},
	    width=0.6\linewidth,
		height=0.4\linewidth,
		ymin=0, ymax=600,
        ytick={150, 300, 450, 600},
		ymajorgrids = true,
		ylabel style={font=\large, at={(-.05, 0.5)}},
		xlabel style={font=\large},
        xticklabel style={rotate=30},
		tick label style={font=\large},
		legend style={font=\large, at={(0.03,0.95)}, anchor=west}, %
		legend cell align={left},
		]
		\addplot [blue!50!white,fill=blue!50!white]
        coordinates {
		(1, 39.4) (2, 78.8) (4, 157.6) (8, 315.2)};
  		\addplot [green!80!white,fill=green!80!white]
        coordinates {
        (1, 65.04) (2, 88.8) (4, 95.36) (8, 117.76)};
		\addlegendentry{GPU}
   		\addlegendentry{CPU}
	\end{axis}
\end{tikzpicture}%
		}
	}
    \hfill
    \subfloat[
	\textbf{Ours.}
	\label{fig:bottleneck_ours}
	]
    {
		\resizebox{0.48\linewidth}{!}{
		\begin{tikzpicture}
	\begin{axis}[
		ybar=10pt,
		bar width=20pt,
        enlarge x limits=0.2,
  		axis y line*=left,
		axis x line*=bottom,
		ylabel={videos/sec},
        symbolic x coords={1, 2, 4, 8},
		xtick=data,
        nodes near coords,
		nodes near coords align={vertical},
		every node near coord/.append style={font=\large, black, /pgf/number format/.cd,fixed zerofill,precision=0},
	    width=0.6\linewidth,
		height=0.4\linewidth,
		ymin=0, ymax=600,
        ytick={150, 300, 450, 600},
		ymajorgrids = true,
		ylabel style={font=\large, at={(-.05, 0.5)}},
		xlabel style={font=\large},
        xticklabel style={rotate=30},
		tick label style={font=\large},
		legend style={font=\large, at={(0.03,0.95)}, anchor=west}, %
		legend cell align={left},
        ]
		\addplot [blue!50!white,fill=blue!50!white] coordinates {
		(1, 62.8) (2, 125.6) (4, 251.2) (8, 502.4)};
  		\addplot [green!80!white,fill=green!80!white] coordinates {
        (1, 82.40) (2, 170.08) (4, 288.32) (8, 421.76)};
		\addlegendentry{GPU}
   		\addlegendentry{CPU}
	\end{axis}
\end{tikzpicture}%
		}
	}
    
    \caption{\small{\textbf{Training throughput \vs number of GPUs} using the standard training pipeline ({\em left}) and ours ({\em right}).
    In the standard training pipeline, the CPU throughput only doubles from single-GPU to 8-GPU scenario as GPUs starve.
    Our pipeline significantly increases both CPU and GPU throughputs.
    For a fair comparison, we keep a constant batch size.
    }}
    \lblfig{bottleneck}
\end{figure*}

\section{Preliminary}
\label{sec:prel}

Let $\vx\in\mathbb{R}^{3 \times T \times H \times W }$ be a video clip of length $T$ and resolution $W \times H$.
The goal of a video model is to analyze this clip and produce a $d$-dimensional output feature $\vy\in\mathbb{R}^{d}$.
This feature may correspond to an embedding in large vision-language models~\cite{lin2022egovlp}, a classifier in action recognition~\cite{feichtenhofer2019slowfast}, or a generic feature map for down-stream applications~\cite{wang2022internvideo}.

\myparagraph{Video Transformer.}
We focus much of our modeling assumptions and improvements on the joint space-time Vision Transformer (ViT)~\cite{dosovitskiy2021vit}.
For any video clip $\vx\in\mathbb{R}^{3 \times T \times H \times W }$, we first divide it into $ N = \frac{T}{t} \times \frac{H}{h} \times \frac{W}{w} $ non-overlapping cubes of size $ t \times h \times w $.
For each cube, the ViT learns a visual token with $ D $ channels, and a space-time positional embedding in the form of learnable parameters $ \mathrm{PE}\in\mathbb{R}^{N \times D} $.
Each visual token then passes through $ L $ Transformer Encoder layers, each of which contains a multi-head self-attention (MHA) layer and a 2-layer MLP.
As we will show in the next section, the ViT is an ideal candidate for large-batch training.
With minor architectural improvements, the ViT is more memory efficient than more complex baselines~\cite{liu2021swin,liu2022videoswin,fan2021mvit,li2022mvitv2,bertasius2021timesformer}.
At the same time, it is more than capable of reaching a state-of-the-art performance on large video-language tasks.

\myparagraph{Flash Attention.}
Attention~\cite{vaswani2017attention} computes a weighted average of the input features, whose weights are determined by the dot-product similarities between the key and query elements on an input sequence.
For $N$ keys and queries, a na\"ive implementation of attention does not only require $O(N^2)$ computation but also $O(N^2)$ memory.
This memory consumption matters for two reasons: First, it limits the maximum batch size. Second, most operations in attention are memory-bound and thus limit throughput.

FlashAttention~\cite{dao2022flashattention} resolves the memory bottleneck of the attention operations.
First, it computes softmax weights implicitly, shrinking the overall memory footprint to $O(N)$.
Second, it computes attention in a block-wise fashion making use of highly efficient on-chip SRAM caches.

\myparagraph{Video Training Pipeline.}
A typical training pipeline for video models works similarly to that for image models.
First, it reads a video clip as a compressed bitstream and decodes the bitstream into a sequence of frames.
Next, a subset of the frames are randomly selected, grouped into a tensor over time, and passed through a set of transformations, or data augmentations.
Typical augmentations include (1) cropping into a fixed target size,~\eg RandomResizedCrop at training and CenterCrop at validation, (2) geometric augmentations such as Flipping and Rotation, (3) photometric augmentations such as ColorJittering and GaussianBlurring, and (4) normalization.
Finally, the transformed tensors from all videos in the same batch are collated and fed into the video model.
In this pipeline, loading video is an IO-bound operation.
Both decoding and transformations are CPU intensive while the model is executed on the GPU side.

\myparagraph{Video Decoder.}
A video decoder takes as input a compressed video bitstream and performs decompression on it.
Decoding speed is determined by various factors, including (1) the size of the bitstream, (2) an efficient frame-seeking strategy to locate the closest key-frames, and (3) slice- or frame-level multi-threading.

\section{Method}

Training of large video models is bottlenecked on two fronts: memory consumption and throughput.
A model's memory consumption limits the maximum batch size, which in turn reduces throughput and even impacts its convergence rate for embedding-based training~\cite{chen2020simclr,he2020moco,radford2021clip,oord2018infonce}.
In the absence of a dedicated video storage and decoding machine, standard IO and pre-processing pipelines are not able to keep up with the GPU throughput, especially on a multi-GPU node.
\shortreffig{bottleneck} illustrates the impact of these bottlenecks on the training throughput.
We show how to reduce each of these bottlenecks and obtain a video training pipeline that performs up to $9\times$ faster.

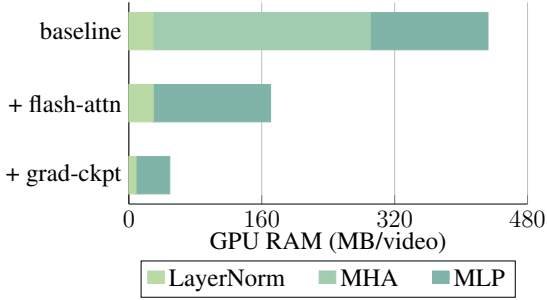
\begin{figure}[!t]
    \centering
    {
		\resizebox{0.9\linewidth}{!}{
        \begin{tikzpicture}[
            every axis/.style={
            width=\linewidth,
            height=.6\linewidth,
            xbar stacked,
            bar width=18pt,
            enlarge y limits=0.2,
            xmin=0, xmax=480,
            axis y line*=left,
            axis x line*=bottom,
            x tick label style = {anchor=north},
            ytick = data, yticklabels = {baseline, + flash-attn, + grad-ckpt},
            xmajorgrids = true,
            xtick={0, 160, 320, 480},
            xlabel style={align=center, font=\large, at={(0.5, 0.05)}},
            xlabel={GPU RAM (MB/video)},
            ylabel style={font=\large},
            tick label style={font=\large},
            legend columns=2,
            legend style={legend columns=3, font=\large, at={(0.5, -0.28)}, anchor=north},
            legend cell align={left},
            },    
        ]
        \begin{axis}[legend style={/tikz/every even column/.append style={column sep=0.5cm}}]
            \addplot [green!80!white,fill=green!80!white] coordinates {
        		(29.04, 2) (29.04, 1) (8.38, 0)};
        		\addplot [darkgreen!80!white,fill=darkgreen!80!white] coordinates {
        		(261.80, 2) (0.085, 1) (0.025, 0)};
        		\addplot [darkergreen!80!white,fill=darkergreen!80!white] coordinates {
        		(141.55, 2) (141.55, 1) (40.87, 0)};
                \legend{LayerNorm, MHA, MLP}
            \end{axis}
        \end{tikzpicture}
        }
	}
    \vspace{-5pt}
    \caption{
    \small{
    \textbf{Memory footprint of the Video ViT}
    for an input clip of resolution $224 \times 224$ and $4$ frames, and cube size $16 \times 16 \times 1$ without a temporal extent.
    Longer clips exhibit a similar memory footprint.
    We consider three variants: a plain ViT baseline, a ViT with FlashAttention~\cite{dao2022flashattention}, and a ViT with FlashAttention and gradient checkpointing~\cite{chen2016gradientcheckpointing}.
    The ViT features three layers that consume memory: LayerNorm, Multi-Head Attention (MHA), and Multi-Layer Perceptrons (MLP).
    }}
    \lblfig{memory_profile}
\end{figure}

\subsection{A Memory-Efficient Video ViT}
\label{sec:method:model}

\shortreffig{memory_profile} analyzes the overall memory consumption of the video ViT model.
In a plain video ViT, the attention operator dominates the overall memory consumption with $ >60\% $ of the memory use.
We completely remove this memory footprint through the use of FlashAttention~\cite{dao2022flashattention}.
We can further trade computation for memory efficiency through gradient checkpointing~\cite{chen2016gradientcheckpointing}.
Due to the isotropic nature of Vision Transformer, where the output shape after each layer is identical throughout the network, the memory complexity can be reduced from $ O(LND) $ to $ O(\sqrt{L}ND) $ for $L$ layers, of $N$ tokens of dimension $D$.

\begin{figure}[!tb]
    \centering
    \subfloat[
	\small{\textbf{Maximum batch size.}}
	\label{fig:max_bs}
	]
    {
		\resizebox{0.48\linewidth}{!}{
		\begin{tikzpicture}
	\begin{axis}[
		ybar=10pt,
		bar width=30pt,
		enlargelimits=0.2,
		axis y line*=left,
		axis x line*=bottom,
		ylabel={\# of videos},
		symbolic x coords={baseline, + flash-attn, + grad-ckpt},
		xtick=data,
		nodes near coords,
		nodes near coords align={vertical},
		every node near coord/.append style={font=\Huge, black, /pgf/number format/.cd,fixed zerofill,precision=0},
        point meta=rawy,
		width=1.5\linewidth,
		height=1.2\linewidth,
		ymin=10, ymax=1e3,
        ymode=log, log origin=infty,
		ymajorgrids = true,
		ylabel style={font=\Huge, at={(-.05, 0.5)}},
		xlabel style={font=\Huge},
        xticklabel style={rotate=30},
		tick label style={font=\Huge},
		legend style={font=\Huge, at={(0.03,0.95)}, anchor=west}, %
		legend cell align={left},
		]
		\addlegendimage{empty legend}
		\addplot [green!80!white,fill=green!80!white] coordinates {
		(baseline, 28) (+ flash-attn, 31) (+ grad-ckpt, 180)};
		\addplot [blue!50!white,fill=blue!50!white] coordinates {
		(baseline, 22) (+ flash-attn, 68) (+ grad-ckpt, 304)};
		\addlegendentry{\hspace{-0.2cm}Video Architecture}
		\addlegendentry{TSF-B}
   		\addlegendentry{ViT-B}
	\end{axis}
\end{tikzpicture}%
		}
	}
    \subfloat[
	\small{\textbf{Training throughput.}}
	\label{fig:throughput}
	]
    {
		\resizebox{0.48\linewidth}{!}{
		\begin{tikzpicture}
	\begin{axis}[
		ybar=10pt,
		bar width=30pt,
		enlargelimits=0.2,
		axis y line*=left,
		axis x line*=bottom,
		ylabel={videos/sec},
		symbolic x coords={baseline, + flash-attn, + grad-ckpt},
		xtick=data,
		nodes near coords,
		nodes near coords align={vertical},
		every node near coord/.append style={font=\Huge, black, /pgf/number format/.cd,fixed zerofill,precision=1},
		width=1.5\linewidth,
		height=1.2\linewidth,
		ymin=30, ymax=100,
		ymajorgrids = true,
		ylabel style={font=\Huge, at={(-.05, 0.5)}},
		xlabel style={font=\Huge},
        xticklabel style={rotate=30},
		tick label style={font=\Huge},
		legend style={font=\Huge, at={(0.03,0.95)}, anchor=west}, %
		legend cell align={left},
		]
		\addlegendimage{empty legend}
		\addplot [green!80!white,fill=green!80!white]
        coordinates {
		(baseline, 39.4) (+ flash-attn, 44.8) (+ grad-ckpt, 30.3)};
		\addplot [blue!50!white,fill=blue!50!white] coordinates {
		(baseline, 45.8) (+ flash-attn, 78.1) (+ grad-ckpt, 62.8)};
		\addlegendentry{\hspace{-0.2cm}Video Architecture}
		\addlegendentry{TSF-B}
   		\addlegendentry{ViT-B}
	\end{axis}
\end{tikzpicture}%
		}
	}
    \vspace{-5pt}
    \caption{\small{\textbf{Throughput and maximum batch size} for a video-text Dual-Encoder model~\cite{zhao2022learning} using a TimeSformer-Base (TSF-B) and ViT-Base (ViT-B) architecture.
    We use $4$ input frames.
    The numbers are measured on a single A5000 (24GB) GPU using \texttt{torch.float16}.
    All input data is kept on the GPU memory for benchmarking purposes only.}}
    \label{fig:memory_usage}
\end{figure}
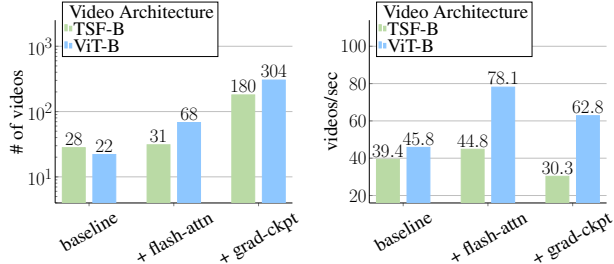

\myparagraph{Discussion.}
With sufficient memory optimization, the plain Video ViT is a very memory and computationally-efficient video architecture.
Due to the efficient block-wise GPU accelerated implementation of FlashAttention the potential cubic computational cost of attention in a Video ViT is not an issue for the workloads to experimented with.
(1) Compared to anisotropic (or hierarchical) Transformer architectures,~\eg Swin~\cite{liu2021swin,liu2022videoswin} or MViT~\cite{fan2021mvit,li2022mvitv2}, ViT contains fewer memory-bound operations, such as window-shifting and pooling.
(2) Compared to another isotropic architecture TimeSformer~\cite{bertasius2021timesformer}, which reduces FLOPs by conducting spatial-only or temporal-only attention separately, ViT has a smaller per-sample memory footprint with gradient checkpointing since the model parameters and number of attention layers are halved.
We illustrate this effect in \shortreffig{memory_usage}.
A memory-efficient ViT with FlashAttention achieves $1.7\times$ throughput than the baseline and $ 3\times $ batch size.
Gradient checkpointing increases the batch size by $ 13.8 \times $, at the cost of slightly reduced throughput ($ 1.4\times $).
(3) Finally, the ViT benefits from large-scale pre-trained image models on vision-language tasks~\cite{radford2021clip} or self-supervised objectives~\cite{he2022mae}.
Starting from pre-trained image models significantly speeds up training on videos.

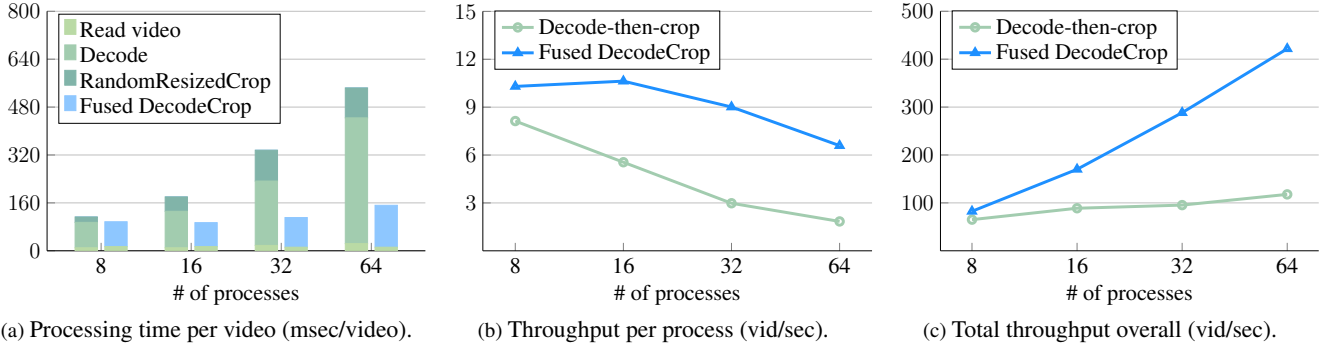
\begin{figure*}[!tb]
    \centering
    \subfloat[
	\small{Processing time per video (msec/video).}
	\label{fig:transform_speed_a}
	]
    {
		\resizebox{0.33\linewidth}{!}{
		\begin{tikzpicture}[
    every axis/.style={
    width=.5\linewidth,
    height=.35\linewidth,
    ybar stacked,
    bar width=12pt,
    enlarge x limits=0.2,
    ymin=0, ymax=800,
    axis y line*=left,
    axis x line*=bottom,
    xlabel={\# of processes},
    x tick label style = {anchor=north},
    symbolic x coords={8, 16, 32, 64},
    xtick=data,
    ymajorgrids = true,
    ytick={0, 160, 320, 480, 640, 800},
    ylabel style={align=center, font=\large, at={(-.08, 0.5)}},
    xlabel style={font=\large},
    tick label style={font=\large},
    legend columns=1,
    legend style={font=\large, at={(0.03,0.76)}, anchor=west},
    legend cell align={left},
    },    
]
	\begin{axis}[bar shift=-8pt,]
		\addplot [green!80!white,fill=green!80!white] coordinates {
		(8, 10) (16, 10) (32, 18) (64, 24)};
		\addplot [darkgreen!80!white,fill=darkgreen!80!white] coordinates {
		(8, 85) (16, 122) (32, 215) (64, 420)};
        \addplot [darkergreen!80!white, fill=darkergreen!80!white] coordinates {
		(8, 18) (16, 48) (32, 103) (64, 100)};
		\addplot [blue!50!white,fill=blue!50!white] coordinates {
		(8, 0) (16, 0) (32, 0) (64, 0)};
		\addlegendentry{Read video}
		\addlegendentry{Decode}
		\addlegendentry{RandomResizedCrop}
   		\addlegendentry{Fused DecodeCrop}
    \end{axis}
    
	\begin{axis}[bar shift=8pt, hide axis]
        \addplot+[green!80!white,fill=green!80!white] 
        coordinates {(8, 14) (16, 14) (32, 12) (64, 12)};
		\addplot+[blue!50!white,fill=blue!50!white] coordinates {
		(8, 83) (16, 80) (32, 99) (64, 140)};
    \end{axis}
\end{tikzpicture}%
        }
    }
    \subfloat[
	\small{Throughput per process (vid/sec).}
	\label{fig:transform_throughput_per_process}
	]
    {
		\resizebox{0.33\linewidth}{!}{
		\begin{tikzpicture}
	\begin{axis} [
		axis x line*=bottom,
		axis y line*=left,
        width=.512\linewidth,
        height=.35\linewidth,
        legend pos=north east,
        ymin=0, ymax=15,
        xlabel={\# of processes},
        x tick label style = {anchor=north},
        symbolic x coords={8, 16, 32, 64},
        xtick=data,
        ymajorgrids = true,
        ytick={3, 6, 9, 12, 15},
        ylabel style={align=center, font=\large, at={(-.08, 0.5)}},
        xlabel style={font=\large},
        tick label style={font=\large},
        legend style={font=\large, at={(0.03,0.88)}, anchor=west},
        legend cell align={left},
    ]

		\addplot[mark=o,darkgreen!80!white,style={ultra thick}] plot coordinates {
			(8, 8.13)
            (16, 5.55)
            (32, 2.98)
            (64, 1.84)
		};
		\addplot[mark=triangle,blue,style={ultra thick}] plot coordinates {
            (8, 10.30)
            (16, 10.63)
            (32, 9.01)
            (64, 6.59)
		};
		\addlegendentry{Decode-then-crop}
		\addlegendentry{Fused DecodeCrop}
 \end{axis}
\end{tikzpicture}%
        }
    }
    \subfloat[
	\small{Total throughput overall (vid/sec).}
	\label{fig:transform_throughput}
	]
    {
		\resizebox{0.33\linewidth}{!}{
		\begin{tikzpicture}
	\begin{axis} [
		axis x line*=bottom,
		axis y line*=left,
		legend pos=north east,
        ymin=0, ymax=500,
        width=.5\linewidth,
        height=.35\linewidth,
        xlabel={\# of processes},
        x tick label style = {anchor=north},
        symbolic x coords={8, 16, 32, 64},
        xtick=data,
        ymajorgrids = true,
        ytick={100, 200, 300, 400, 500},
        ylabel style={align=center, font=\large, at={(-.08, 0.5)}},
        xlabel style={font=\large},
        tick label style={font=\large},
        legend style={font=\large, at={(0.03,0.88)}, anchor=west},
        legend cell align={left},
    ]

		\addplot[mark=o,darkgreen!80!white,style={ultra thick}] plot coordinates {
			(8, 65.04)
            (16, 88.8)
            (32, 95.36)
            (64, 117.76)
		};
		\addplot[mark=triangle,blue,style={ultra thick}] plot coordinates {
            (8, 82.40)
            (16, 170.08)
            (32, 288.32)
            (64, 421.76)
		};
		\addlegendentry{Decode-then-crop}
		\addlegendentry{Fused DecodeCrop}
 \end{axis}
\end{tikzpicture}%
        }
    }
    
    \caption{\small{
    \textbf{CPU utilization of a standard video processing pipeline \vs ours.}
    We build an in-memory toy dataset of 1,024 15-second video clips and measure the average elapsed time of sampling $4$ frames with a pool of $ M $ processes, where $ M $ varies across $ \{ 8, 16, 32, 64 \} $.
    We measure (a) the processing time per video (latency), (b) the throughput per process, and (c) the overall throughput of the video loader.
    The numbers are measured on a server with $ 2\times $ Intel Xeon 24-Core CPU @ 2.20GHz (96 threads in total). We ignore other augmentation techniques in this experiment.}
    }
    \lblfig{transform_speed}
\end{figure*}

\subsection{Increasing CPU Utilization in Pre-processing}
\label{sec:method:transform}

With a larger batch size, video pre-processing becomes a clear bottleneck.
Without dedicated hardware solutions, the CPU on a single node server is simply not able to supply eight GPUs with sufficient data, and thus largely starves the GPUs.
This effect is highlighted in \shortreffig{transform_speed}.
At its peak, a video ViT is able to process $60\unsim70$ video clips per second per GPU or $400\unsim500$ clips per second on an 8-GPU node.
A standard video training pipeline supplies at most $100\unsim120$ clips per second, thus utilizing GPUs at $\unsim25\%$.
Increasing the number of worker threads only marginally improves the pipeline efficiency.

As shown in \shortreffig{transform_speed_a}, a standard video pipeline spends the majority of its computation on decoding and the random resized cropping (RRC) operation.
It first completely decodes a larger-than-needed video clip, and subsequently crops it, both of which are CPU and CPU-memory intensive operations.
To address this, we propose to merge RRC into the video decoding stage as a cropping filter.

\myparagraph{RandomResizedCrop (RRC).}
RRC~\cite{szegedy2015inception} takes as input three tuples, namely the target size $ (H_t, W_t) $,  scale range $ (s_{min}, s_{max}) $, and aspect ratio range $ (r_{min}, r_{max}) $. 
First, it computes the area of the frame $ (HW) $.
Second, it randomly sample a target area $ A $ and aspect ratio $ r $ by $ A \sim U(s_{min}HW, s_{max}HW), r \sim U(r_{min}, r_{max}) $  so that the cropping size should be:
\begin{align}
    W_{crop} = \lfloor \sqrt{Ar} \rceil, H_{crop} = \lfloor \sqrt{A/r} \rceil
\end{align}
Next, it randomly samples the left edge and the top edge:
\begin{align}
    x = \lfloor U(0, W - W_{crop}) \rceil, y = \lfloor U(0, H - H_{crop}) \rceil.
\end{align}
Finally, the cropped output $ \vx[:, :, y:y+H_{crop}, x:x+W_{crop}] $ is rescaled to $ \vx' \in \mathbb{R}^{T\times 3 \times H_t \times W_t} $.

\myparagraph{RandomResizedCrop as a cropping filter.}
The cropping region is only conditioned on the frame size $ (H, W) $ and agnostic to the frame contents.
We thus first generate cropping coordinates from the meta-data, specifically the width and height, of the video without decoding it.
Next, we conduct decoding and cropping simultaneously by adding a cropping filter at the video decoder.
This ensures that the video decoder executes the minimal amount of decoding operations to retrieve the data needed.
\shortreffig{decode_and_rrc} in \shortrefsec{supp:pseudo_code} illustrates the Pythonic pseudo-code.
The resulting data-loader features a close to linear scaling as the process pool increases from 8 to 64 processes (\shortreffig{transform_throughput}).
The latency only increases from 97 to 152ms per video per process (\shortreffig{transform_speed_a}).

\myparagraph{Beyond RandomResizedCrop.}
Fused DecodeCrop naturally extends to most cropping-based augmentation,~\eg SimpleRandomCropping, which was first proposed in Alex-Net~\cite{krizhevsky2012imagenet} and recently reused in DeiT III~\cite{touvron2022deit3} to great effect.

After cropping, all tensors have a fixed shape and are readily batched.
We move the data to the GPU at this stage and apply other augmentations, such as photometric augmentation and normalization, on the GPU.
This eliminates the CPU bottleneck in current video training pipelines.

The final bottleneck is disk IO, as most video datasets are too large to fit into memory.

\subsection{Eliminating IO bottleneck for Long Videos}
\label{sec:method:io}

Long-term videos have become an increasingly popular resource for multi-modal contrastive pre-training~\cite{alayrac2020self,miech2020milnce}.
The most straightforward way is to trim the long videos according to the annotated timestamps beforehand.
The drawbacks are twofold: (1) Trimming may increase the storage if there are multiple annotations in one video and the annotated clips overlap.
(2) Trimming ignores the large proportion of the unannotated parts, which may have benefited video representation through pseudo-labeling~\cite{zhao2022learning}.

An attractive alternative way is to split each input video into multiple fixed-length chunks~\cite{lin2022egovlp,zhao2022learning}.
The length of these chunks is often chosen heuristically,~\eg $T = 5\unsim10~\mathrm{min}$ long.
The trade-offs are clear: Shorter chunks reduce the IO bottleneck. Longer chunks reduce potential duplication of the input data.
Ideally, one chooses the largest chunk size that reduces the IO bottleneck.

Let $B$ denote the batch size, $\rho$ denote the average bitrate of a video, $S_r$ denote the maximum read speed, and $\Delta$ denote the time of a training step.
To hide the IO bottleneck from the training, we require the video model to consumer fewer bits $B \times \rho \times T$ than the disk can afford $S_r \times \Delta$:
\begin{align}
    B \times \rho \times T \leq S_{r} \times \Delta.
    \label{eq:io}
\end{align}
Note, that we only control the length $T$ of each chunk.
The bitrate $ \rho $ depends on the resolution and the codec.
Maximum read speed $S_{r}$ varies significantly according to the hardware,~\eg $80~\mathrm{MB/sec}$ for HDD, $500~\mathrm{MB/sec}$ for SATA SSD, and $3~\mathrm{GB/sec}$ for NVMe SSD.
In our experimental setup, typical values are $N=1024$, $\rho=1~\mathrm{Mb/sec}$, $\Delta=4~\mathrm{sec}$ and $S_{r}=500~\mathrm{MB/sec}$, which leads to $T\leq 16~\mathrm{sec}$.
We use 15-second chunks in practice to avoid GPU starvation due to fluctuations in the disk read speed.
For most video tasks, the size of the video clip fed into the network is much smaller than our chunk size.
The pipeline thus avoids having to read multiple consecutive chunks.

\section{Experiments}

To show the effectiveness of our expedited training pipeline, we conduct video-language pre-training on the Ego4D egocentric video dataset and evaluate the performance on Epic-Kitchens 100 (EK-100).
We summarize dataset statistics and evaluation protocols in \shortrefsec{expts:datasets}.
Experimental setups including the model configuration and the hardware specifications are elaborated in \shortrefsec{expts:setup}. 
After discussing the main results in \shortrefsec{expts:main} and ablation studies in \shortrefsec{expts:ablation}, we present an application of our optimizing techniques to other representative video models in \shortrefsec{expts:videomae}. 

\subsection{Datasets and Evaluation Protocols}
\label{sec:expts:datasets}

\myparagraph{Ego4D}~\cite{grauman2022ego4d} is the largest egocentric video dataset to date, including 3,670 hours of videos with temporally dense free-form narratives.
Following the training split and pairing strategy in EgoVLP~\cite{lin2022egovlp}, we get around 4M video-text pairs with an average length of 1 second.
These pairs are further augmented by LaViLa~\cite{zhao2022learning} to boost contrastive pre-training.

\myparagraph{EK-100}~\cite{damen2022epickitchen} is a popular and challenging egocentric video recognition benchmark with 100 hours of cooking scenarios.
We focus on two tasks: Multi-Instance Retrieval (\textbf{EK-100 MIR}) and Action Recognition (\textbf{EK-100 CLS}).
The MIR task requires retrieving the text given videos (V$\rightarrow$T) and videos given text (T$\rightarrow$V).
It contains 67,217/9,668 video-text pairs in the training/testing split respectively.
We use two evaluation protocols: (1) \emph{Zero-shot}, meaning that we apply the video-text encoders pre-trained on Ego4D directly on the EK-100 testing split without any additional tuning; (2) \emph{Fine-tuned}, meaning that we take the pre-trained video-text encoder and perform end-to-end fine-tuning on the EK-100 training split.
The evaluation metrics are mean Average Precision (mAP) and normalized Discounted Cumulative Gain (nDCG) of V $\rightarrow$ T, T $\rightarrow$ V, as well as the average of V $\rightarrow$ T and T $\rightarrow$ V.
The CLS task requires classifying each video clip into one of 97 verbs and 300 nouns each, resulting in a combination of 3,806 action categories.
We report top-1 accuracy on verbs, nouns, and actions after finetuning the video encoder. Among the three accuracies, the action-level accuracy is emphasized.

\subsection{Experimental Setups}
\label{sec:expts:setup}

\myparagraph{Video-language model architecture.}
The video-language model follows CLIP~\cite{radford2021clip}, which is composed of a vision encoder and a text encoder.
The vision encoder is a Vision Transformer Base (ViT-B) model, whose weights are initialized from CLIP~\cite{radford2021clip} except that we randomly initialize the temporal position embedding $ \mathrm{PE}_t \in \mathrm{R}^{T\times N\times D} $ and add it to the original spatial position embedding $ \mathrm{PE}_s \in \mathrm{R}^{N\times D} $,~\ie $ \mathrm{PE}[i,:,:] = \mathrm{PE}_t[i,:,:] + \mathrm{PE}_s $.
We represent each video clip by $ T=4 $ frames when pre-training on Ego4D.
When fine-tuning on EK-100, we increase $ T $ from 4 to 16 and linearly interpolate $ \mathrm{PE}_t $ along the temporal dimension.
The text encoder is a 12-layer GPT-like Transformer~\cite{radford2019gpt2,vaswani2017attention}.
It takes as input one video narrative after a BPE tokenizer~\cite{sennrich2016bpe} with at most 77 tokens.
With memory-efficient attention, gradient checkpointing, and automatic mixed-precision training, we are able to fit $ 256 $ video clips on a 24GB GPU so that the total batch size will be 2,048.

\myparagraph{Hardware.}
We conduct experiments on two types of hardware. 
One is a server with $ 8\times $ NVIDIA RTX A5000 GPU and $ 2\times $  Intel Xeon Gold 5220R 24-Core CPU @ 2.20GHz (96 threads in total); the other has $ 4\times $ A5000 GPU and $ 1\times $ AMD Ryzen Threadripper PRO 5975WX 32-Core CPU (64 threads).
The videos reside on an NVMe data server via a Network File System (NFS) with 10Gbps Ethernet.
Both of the hardware is much more available in academia compared to a gigantic cluster of A100 GPUs inter-connected by InfiniBand. 
We report the main quantitative results by using the 8-GPU server and perform the stress test on data loading using the 4-GPU one if not otherwise specified.

\subsection{Main Results}
\label{sec:expts:main}

We present our main results from two aspects: training efficiency compared to previous works on Ego4D, and strong accuracy \vs prior methods on EK-100 MIR.

\begin{table*}[!tb]
    \centering
    \tablestyle{3pt}{1.05}
    \begin{tabular}{c|c|c|c|c|c|c|c}
    Method & Corpus size & Hardware & Batch size & Memory & GPU$\cdot$hour & kg CO$_2$eq. & 0-shot Avg. mAP \\
    \thickhline
    \multicolumn{8}{l}{\textsc{(Original narratives)}} \\
    \hline
    EgoVLP~\cite{lin2022egovlp} & 3.8M & $ 32\times $ A100 & 16 & 22 & 1,536 & 227.33 & 23.3 \\
    Ours & 4.0M & $ 8\times $ A5000 & 256 & 19 & 130~\gain{(-92\%)} & 11.06~\gain{(-94\%)} & 28.4~\gain{(+5.1)} \\
    \hline
    \multicolumn{8}{l}{\textsc{(LLM-augmented)}} \\
    \hline
    LaViLa~\cite{zhao2022learning} & 35.0M & $ 32\times $ V100 & 32 & 25 & 1,824 & 202.46 & 30.9 \\
    Ours & 35.0M & $ 8\times $ A5000 & 256 & 19 & 260~\gain{(-86\%)} & 22.12~\gain{(-89\%)} & 32.7~\gain{(+1.8)} \\
    \hline
    \end{tabular}
    
    \caption{
    \small{
    \textbf{Pre-training cost and 0-shot generalization performance of large video-language models} on EK-100 MIR.
    We compare our training pipeline to the standard training pipeline for large video-language models for two baselines: EgoVLP~\cite{lin2022egovlp} and LaViLa~\cite{zhao2022learning}.
    Each baseline was originally trained on a multi-node cluster, while our training pipeline fits onto a single 8-GPU machine.
    We compare training time (GPU$\cdot$hours), total carbon emission (kg CO$_2$eq.) estimated using~\cite{lacoste2019quantifying,mlco2}, and zero-shot generalization performance to EK-100 MIR.
    }}
    \label{tab:cost}
\end{table*}

\myparagraph{Pre-training efficiency on Ego4D.}
We compare the compute cost for Ego4D pre-training in~\Cref{tab:cost}.
With the original 4M ground-truth narratives, our model can be trained in 5 full epochs using $ 8\times $ A5000 GPUs in 18 hours. 
In contrast, it takes 1,536 GPU$\cdot$hours to train an EgoVLP~\cite{lin2022egovlp} video-text model, which is around $ 11.8\times $ than ours.
Thanks to the increased batch size, the zero-shot result is also better: ours is 4.7\% better than EgoVLP in terms of zero-shot average mAP on EK-100 MIR.
The effect of batch size on embedding losses is generally well understood, and higher batch sizes almost always lead to better performance~\cite{radford2021clip}.

Our pipeline also benefits from larger-scale video-text pairs generated by Visual Language Models~\cite{zhao2022learning}.
We follow LaViLa~\cite{zhao2022learning} and extend the training schedule to cover 10 ``effective" epochs.
In this setting, our training pipeline achieves an mAP of 31.7\% within 33 hours.
This is 2.2\% higher at $ \frac{1}{5} $ of the compute cost of LaViLa. 
The increase in performance is again likely due to the larger batch size.

\begin{table}[!tb]
    \centering
    \tablestyle{1pt}{1.05}
    \begin{tabular}{c|c|c|c|c|c|c|c}
    \multirow{2}{*}{Method} & \multirow{2}{*}{Backbone} & \multicolumn{3}{c|}{mAP} & \multicolumn{3}{c}{nDCG} \\
    & & V$\rightarrow$T & T$\rightarrow$V & Avg. & V$\rightarrow$T & T$\rightarrow$V & Avg. \\
    \thickhline
    \multicolumn{8}{l}{\textsc{(Zero-shot)}} \\
    \hline
    EgoVLP~\cite{lin2022egovlp} & TSF-B & 19.4 & 13.9 & 16.6 & 24.1 & 22.0 & 23.1 \\
    EgoVLP$^*$~\cite{lin2022egovlp,zhao2022learning} &  TSF-B  & 26.0 & 20.6 & 23.3 & 28.8 & 27.0 & 27.9 \\
    LaViLa~\cite{zhao2022learning} & TSF-B & 35.1 & 26.6 & 30.9 & 33.7 & 30.4 & 32.0 \\
    Ours & ViT-B & \textbf{37.1} & \textbf{28.7} & \textbf{32.9} & \textbf{34.4} & \textbf{31.0} & \textbf{32.7} \\
    \rowcolor{Gray} LaViLa~\cite{zhao2022learning} & TSF-L & 40.0 & 32.2 & 36.1 & 36.1 & 33.2 & 34.6 \\
    \rowcolor{Gray} Ours & ViT-L & \textbf{41.7} & \textbf{33.5} & \textbf{37.6} & \textbf{36.8} & \textbf{33.9} & \textbf{35.3} \\
    \hline
    \multicolumn{8}{l}{\textsc{(Finetuned)}} \\
    \hline
    MME~\cite{wray2019jpose} & TBN & 43.0 & 34.0 & 38.5 & 50.1 & 46.9 & 48.5 \\
    JPoSE~\cite{wray2019jpose} & TBN & 49.9 & 38.1 & 44.0 & 55.5 & 51.6 & 53.5 \\
    EgoVLP~\cite{lin2022egovlp} & TSF-B & 49.9 & 40.5 & 45.0 & 60.9 & 57.9 & 59.4 \\
    LaViLa~\cite{zhao2022learning} & TSF-B & 55.2 & 45.7 & 50.5 & 66.5 & 63.4 & 65.0  \\
    Ours & ViT-B & \textbf{55.9} & \textbf{47.8} & \textbf{51.8} & \textbf{68.2} & \textbf{65.4} & \textbf{66.8} \\
    \rowcolor{Gray} LaViLa~\cite{zhao2022learning} & TSF-L & 54.7 & 47.1 & 50.9 & 68.1 & 64.9 & 66.5 \\
    \rowcolor{Gray} Ours & ViT-L & \textbf{57.9} & \textbf{51.1} & \textbf{54.5} & \textbf{70.4} & \textbf{67.6} & \textbf{69.0} \\
    \hline
    \end{tabular}

    \caption{
    \small{
    \textbf{The performance of multi-instance retrieval on EK-100.}
    Our method outperforms previous works on both zero-shot and fine-tuned settings with similar model complexity.
    Specifically, our model with a ViT-Base video encoder achieves \textbf{2.3\%} higher zero-shot mAP than LaViLa with TimeSformer-Base.
    Note that this is achieved with a significantly smaller amount of compute cost, details of which are given in~\Cref{tab:cost}.
    EgoVLP$^*$ indicates that we evaluate the EgoVLP's checkpoint using our data format for a fair comparison.
    }}
    \label{tab:ek100_mir}
\end{table}

\begin{table}
	\tablestyle{1pt}{1.05}
	\begin{tabular}{c|c|ccc}
		\multirow{2}{*}{Method (Backbone)}&  \multirow{2}{*}{Pretrain Data} & \multicolumn{3}{c}{Top-1 accuracy} \\
		&  & Verb & Noun & Action \\
		\thickhline
		IPL (I3D)~\cite{wang2021ipl} & K400 & 68.6 & 51.2 & 41.0 \\
		ViViT-L~\cite{arnab2021vivit} & IN-21k+K400 & 66.4 & 56.8 & 44.0 \\
		MoViNet~\cite{kondratyuk2021movinets} & N/A & \underline{72.2} & 57.3 & 47.7 \\
		MTV~\cite{yan2022multiview}  & WTS-60M & 69.9 & \underline{63.9} & {50.5} \\
		Omnivore (Swin-B)~\cite{girdhar2022omnivore} & {\scriptsize IN-(21k+1k)+K400+SUN} & 69.5 & 61.7 & 49.9 \\
		MeMViT~\cite{wu2022memvit}  & K600 & 71.4 & 60.3 &  48.4 \\
		LaViLa (TSF-B)~\cite{zhao2022learning} & WIT+Ego4D & {69.0} & {58.4} & {46.9} \\
            \textbf{Ours} (ViT-B) & WIT+Ego4D & {70.0} & {59.8} & {49.1} \\
		\hline
		LaViLa (TSF-L)~\cite{zhao2022learning} & WIT+Ego4D & {72.0} & {62.9} & \underline{51.0} \\
        \textbf{Ours} (ViT-L) & WIT+Ego4D & \textbf{73.0} & \textbf{65.4} & \textbf{54.4} \\
		\hline
	\end{tabular}
        
	\caption{\small{\textbf{The performance of action recognition on EK-100}. We report top-1 accuracy on verbs, nouns, and actions. Ours outperforms all prior works in terms of action-level top-1 accuracy.}}
	\label{tab:ek100_cls}
\end{table}

\myparagraph{EK-100 MIR.}
We evaluate our pre-trained model on EK-100 MIR in~\Cref{tab:ek100_mir} using $ T=16 $ for fair comparison with prior methods.
In the zero-shot setup, our model achieves 33.2\% average mAP and 33.0\% average nDCG, which is 2.3\% and 1.0\% higher than the previous state-of-the-art.
Next, we fine-tuned the video-text encoder on the EK-100 MIR train split by replacing the InfoNCE loss with the max-margin loss following Wray \etal~\cite{wray2019jpose}.
We see a consistent improvement of 1.3\% (51.8 \vs 50.5) in average mAP and 1.8\% (66.8 \vs 65.0) in average nDCG.
When we upgrade the backbone to ViT-Large, the gain is boosted to 3.6\% in average mAP and 2.5\% in average nDCG respectively.

\myparagraph{EK-100 CLS.} We fine-tune our pre-trained model on EK-100 CLS and show results in~\Cref{tab:ek100_cls}.
With ViT-Base as the backbone, our model achieves 49.1\% top-1 action accuracy, which is 2.2\% higher than LaViLa with a similar TimeSformer-Base encoder and the same pre-training data.
It is also comparable with prior methods while requiring significantly fewer pre-training videos.
When we upgrade the backbone to ViT-Large, the gain is amplified: our method achieves 54.4\% top-1 action accuracy, which is 3.4\% higher than LaViLa with TimeSformer-Large. It also beats the best single model from M\&M~\cite{xiong2022mm}, the 2022 EPIC-Kitchens Challenge winner, which uses extra modalities (RGB+Optical Flow+Audio) and doubled resolution (432$\times$432 crop), by a clear margin (53.6\% \vs 54.4\%).

\subsection{Ablation Studies}
\label{sec:expts:ablation}

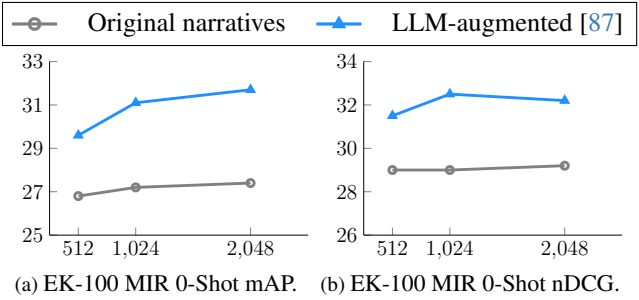
\begin{figure}[!tb]
    \centering
	\begin{subfigure}[b]{\linewidth}
		\centering
		\begin{tikzpicture}
			\begin{customlegend}
				[
				legend columns=2, legend style={column sep=2ex},
				legend entries={Original narratives, LLM-augmented~\cite{zhao2022learning}}
                ]
				\addlegendimage{mark=o,style={very thick},gray}
				\addlegendimage{mark=triangle,style={very thick},blue}
			\end{customlegend}
		\end{tikzpicture}
	\end{subfigure}

    \subfloat[
	\small{EK-100 MIR 0-Shot mAP.}
	\label{fig:ek100_mir_map_vs_bs}
	]
    {
		\resizebox{0.48\linewidth}{!}{
		\begin{tikzpicture}
	\begin{axis} [
		axis x line*=bottom,
		axis y line*=left,
		legend pos=north east,
        ymin=25, ymax=33,
		xmin=256, xmax=2560,
		xticklabel={\pgfmathparse{\tick}\pgfmathprintnumber{\pgfmathresult}},
		xtick={512, 1024, 2048},
		ytick={25, 27, 29, 31, 33},
		width=0.8\linewidth,
		height=0.6\linewidth,
		tick label style={font=\large},
		]

		\addplot[mark=o,style={ultra thick},gray] plot coordinates {
			(512, 26.8)
            (1024, 27.2)
            (2048, 27.4)
		};
		\addplot[mark=triangle,blue,style={ultra thick}] plot coordinates {
            (512, 29.6)
            (1024, 31.1)
            (2048, 31.7)
		};
	\end{axis}
\end{tikzpicture}%
		}
	}
    \subfloat[
	\small{EK-100 MIR 0-Shot nDCG.}
	\label{fig:ek100_mir_ndcg_vs_bs}
	]
    {
		\resizebox{0.48\linewidth}{!}{
		\begin{tikzpicture}
	\begin{axis} [
		axis x line*=bottom,
		axis y line*=left,
		legend pos=north east,
            ymin=26, ymax=34,
		xmin=256, xmax=2560,
		xticklabel={\pgfmathparse{\tick}\pgfmathprintnumber{\pgfmathresult}},
		xtick={512, 1024, 2048},
		ytick={26, 28, 30, 32, 34},
		width=0.8\linewidth,
		height=0.6\linewidth,
		tick label style={font=\large},
		]

		\addplot[mark=o,style={ultra thick},gray] plot coordinates {
			(512, 29.0)
            (1024, 29.0)
            (2048, 29.2)
		};
		\addplot[mark=triangle,style={ultra thick},blue] plot coordinates {
            (512, 31.5)
            (1024, 32.5)
            (2048, 32.2)
		};
	\end{axis}
\end{tikzpicture}%
		}
	}
    \vspace{-5pt}
    \caption{
    \small{
    \textbf{Effect of pre-training batch size.}
    The numbers are reported using $ T= 4 $ frames as input.
    Large-batch training, which was not possible without multi-node training, benefits the video-language contrastive models consistently, especially in the presence of larger-scale narratives. 
    }}
    \label{fig:expts:batch_size}
\end{figure}

\myparagraph{Benefits of large-batch pre-training.}
Next, we further study the benefits of large-batch training for video-language models.
\shortreffig{expts:batch_size} summarizes the results.

First, we observe that a larger corpus size benefits more from the large-batch training:
In the original narratives, any gains are marginal ($ \unsim0.2\% $) with an increased batch size.
However, with additional augmentation by a large language model~\cite{zhao2022learning}, a larger batch size significantly increases the mAP.
One reason might be that the current data scale is still insufficient for training a video-language model in full gear.

Second, with other settings fixed the same, our model with a ViT-Base backbone is consistently better than LaViLa with a TimeSformer-Base backbone.
ViT-Base inherits the full topology of the pre-trained image encoder~\cite{radford2021clip} with the only exception of the temporal positional embedding, making itself easier to fine-tune than TimeSformer.
This reveals the effectiveness of isotropic Transformer architectures compared to other variants given the same setting (~\eg same batch size in our case), echoing the discovery in other tasks~\cite{li2022vitdet,touvron2022deit3}.
Our memory optimization makes it possible to use ViT-Base as is.

\myparagraph{Model runtime after fixing different bottlenecks.}
We analyze the IO and CPU bottleneck separately under simplified conditions in \shortrefsec{method:io} and \shortrefsec{method:transform}.
Here, we measure the runtime of training video-text dual-encoder in the real world.
We summarize our findings in \Cref{tab:runtime_clip} by starting from the LaViLa baseline.
First, shortening the length chunk from 5 minutes to 15 seconds reduces the data-loading overhead by $ 6\times $ and increases the overall training speed by $ 2.6\times $.
Interestingly, this also reduces the model's forward and backward times.
Next, we switch to decoding and cropping simultaneously.
We can see that the data-loading overhead is further reduced by 0.4 seconds per iteration and the overall training speed is faster.

\begin{table}[!tb]
    \centering
    \tablestyle{1pt}{1.05}
    \begin{tabular}{c|c|c|c|c|c|c}
        Batch & Mem.-eff. & Shorter & Merged & Data-loading & Training & Actual \\
        size & Attention & Chunks & RRC & overhead & speed & Throughput \\
        & (\shortrefsec{method:model}) & (\shortrefsec{method:io}) & (\shortrefsec{method:transform}) & (sec/it) & (sec/it) & (vid/sec) \\
        \thickhline
        64 &  &  &  & 0.5 & 3.9 & 130 \\
        64 &  &  & \checkmark & 0.3 & 3.5 & 146 \\
        64 &  & \checkmark &  & 0.1 & 1.84 & 278 \\
        64 &  & \checkmark & \checkmark & 0.1 & 1.84 & 278 \\
        \hline
        256 &  &  &  & (OOM) & (OOM) & N/A \\
        256 & \checkmark &  &  & 10.1 & 20.8 & 98 \\
        256 & \checkmark &  & \checkmark & 8.3 & 17.8 & 115 \\
        256 & \checkmark & \checkmark &  & 1.3 & 6.5 & 315 \\
        256 & \checkmark & \checkmark & \checkmark & 0.9 & 5.9 & 347 \\
        \hline
    \end{tabular}
    \caption{\small{\textbf{The effect on the runtime after improvements to the standard video training pipeline.}
    The original model did not fit in the GPU memory in our setup, while all other improvements significantly reduced the training time.
    }}
    \label{tab:runtime_clip}
\end{table}

\subsection{Application: Expedite Training MAE in Videos}
\label{sec:expts:videomae}

The optimized CPU and GPU computation is not limited to training large video-language models.
We take VideoMAE~\cite{tong2022videomae} as an example.
VideoMAE operates on only a small subset of input tokens and
masks out others.
This leads to light-weight encoder and decoder computations where data-loading becomes a bottleneck~\cite{tong2022videomae,feichtenhofer2022maest,girdhar2022omnimae}.

We conduct VideoMAE pre-training on the training split of Kinetics-400~\cite{carreira2017i3d}, which contains 241,258 videos.
We follow the default setting in~\cite{tong2022videomae}.
The encoder is a standard ViT-Base model while the decoder has 4 additional Transformer Encoder layers.
Each input clip contains 16 frames with a sample stride of 4 and is split into non-overlapping $ 8 \times 14 \times 14 = 1568 $ cubes of size $ t\times h\times w = 2\times 16 \times 16 $.
Since the number of visible tokens at the encoder side is only 10\%, the memory reduction of using memory-efficient attention is marginal.
As such, we only apply memory-efficient attention to the decoder.

\begin{figure}[!tb]
    \centering
	\begin{subfigure}[b]{\linewidth}
		\centering
		\begin{tikzpicture}
			\begin{customlegend}
				[
				legend columns=2, legend style={column sep=2ex},
				legend entries={decode-then-crop, Fused DecodeCrop}
                ]
				\addlegendimage{mark=o,style={very thick},gray}
				\addlegendimage{mark=triangle,style={very thick},blue}
			\end{customlegend}
		\end{tikzpicture}
	\end{subfigure}

    \subfloat[
	\small{Data loading overhead.}
	\label{fig:overhead_vs_num_cpu}
	]
    {
        \resizebox{0.48\linewidth}{!}{
        \begin{tikzpicture}
	\begin{axis} [
		axis x line*=bottom,
		axis y line*=left,
		legend pos=north east,
        ymin=0, ymax=3.2,
		xmin=0, xmax=20,
		xticklabel={\pgfmathparse{\tick}\pgfmathprintnumber{\pgfmathresult}},
		xtick={4, 8, 12, 16},
		ytick={0.8, 1.6, 2.4, 3.2},
        ylabel={(sec/iter)},
        xlabel={\# of workers per GPU},
        ylabel style={align=center, font=\Large, at={(0.02, 0.5)}},
        xlabel style={font=\Large},
        width=0.8\linewidth,
		height=0.6\linewidth,
  		tick label style={font=\Large},
		]
		\addplot[mark=o,style={ultra thick},gray] plot coordinates {
            (4, 2.41)
            (8, 1.45)
            (12, 1.14)
            (16, 0.74)
		};
		\addplot[mark=triangle,blue,style={ultra thick}] plot coordinates {
            (4, 1.82)
            (8, 0.75)
            (12, 0.44)
            (16, 0.25)
		};
	\end{axis}
\end{tikzpicture}%
        }
    }
    \subfloat[
	\small{Training speed.}
	\label{fig:speed_vs_num_cpu}
	]
    {
        \resizebox{0.48\linewidth}{!}{
        \begin{tikzpicture}
	\begin{axis} [
		axis x line*=bottom,
		axis y line*=left,
		legend pos=north east,
            ymin=1, ymax=8,
		xmin=0, xmax=20,
		xticklabel={\pgfmathparse{\tick}\pgfmathprintnumber{\pgfmathresult}},
            y tick label style={
                /pgf/number format/.cd,
                fixed,
                fixed zerofill,
                precision=1
            },
            xtick={4, 8, 12, 16},
		ytick={2.0, 4, 6, 8},
        ylabel={(sec/iter)},
        xlabel={\# of workers per GPU},
        ylabel style={align=center, font=\Large, at={(0.02, 0.5)}},
        xlabel style={font=\Large},
        width=0.8\linewidth,
		height=0.6\linewidth,
  		tick label style={font=\Large},
		]
		\addplot[mark=o,style={ultra thick},gray] plot coordinates {
            (4, 6.46)
            (8, 3.40)
            (12, 2.70)
            (16, 2.40)
		};
		\addplot[mark=triangle,blue,style={ultra thick}] plot coordinates {
            (4, 5.19)
            (8, 2.59)
            (12, 1.90)
            (16, 1.55)
		};
	\end{axis}
\end{tikzpicture}%
        }
    }
    \caption{
    \small{
    \textbf{Training speed comparison of a video MAE model}
    on $ 4\times $ A5000 GPUs and $ 1\times $ AMD 32-Core CPU (64 threads).
    Our Fused DecodeCrop consistently reduces data loading overhead and increases the overall training speed compared to baseline training pipelines.
    }}
    \label{fig:videomae_speed}
\end{figure}
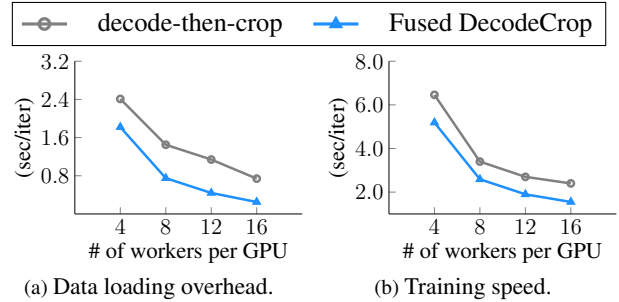

\shortreffig{videomae_speed} shows the improved training speed of using Fused DecodeCrop: It reduces the data loading overhead by almost $ 3\times $,~\ie from 0.74 to 0.25 seconds per iteration.
As a result, the overall training speed decreases from 2.4 to 1.55 seconds per iteration, resulting in a 35\% reduction in training time.
Finally, we conduct a system-level comparison between the original VideoMAE and ours in~\Cref{tab:mae_details} with the same 4-GPU hardware.
Under the same 800-epoch schedule, our training pipeline achieves the same level of accuracy after supervised fine-tuning while running $1.7\times$ faster than VideoMAE. 

\begin{table}[!tb]
    \centering
	\tablestyle{1pt}{1.05}
	\begin{tabular}{c|c|c|c||c}
		Method & backbone & epochs & GPU$\cdot$hour & Top-1/5 Acc. (ft.) \\
		\thickhline
		VideoMAE~\cite{tong2022videomae} & ViT-B & 800 & 995 & 80.0/94.4 \\
		Ours & ViT-B & 800 & 583~\gain{(-41\%)} & 80.0/94.5 \\
		\hline
	\end{tabular}
	\vspace{-5pt}
    \caption{
		\small{\textbf{System-level comparison of training Video MAE.} Both GPU$\cdot$hours are measured on the 4-GPU hardware. Our pipeline achieves the same accuracy after fine-tuning (``ft.'') while using 41\% less pre-training time than VideoMAE~\cite{tong2022videomae}.
    }}
	\label{tab:mae_details}
\end{table}

\section{Conclusion}
We study the bottleneck of training video models from the perspectives of IO, CPU, and GPU computation.
With a combination of a memory-efficient attention-based video model, fused decode-cropping operator, and chunk-based video loading, we show the feasibility of training a state-of-the-art video model in a day on a single machine.

\myparagraph{Acknowledgements.}
This material is based upon work in part supported by the National
Science Foundation under Grant No. IIS-1845485.
YZ would like to thank Lingfan Yu for the helpful discussions on profiling training throughput.

\clearpage

\appendix

\section{Pseudo-code for Fused DecodeCrop}
\label{sec:supp:pseudo_code}

\begin{figure}[tb!]
    \centering
    \begin{subfigure}[b]{\linewidth}
        \includegraphics[width=\textwidth]{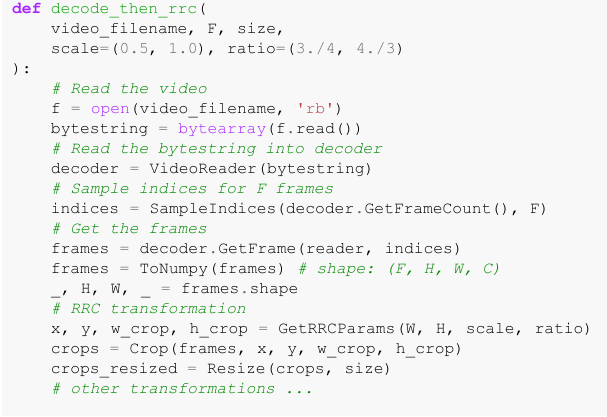}
    \caption{\small{Decode-then-crop.}}
	\label{fig:decode_then_rrc}
    \end{subfigure}
    \begin{subfigure}[b]{\linewidth}
        \includegraphics[width=\textwidth]{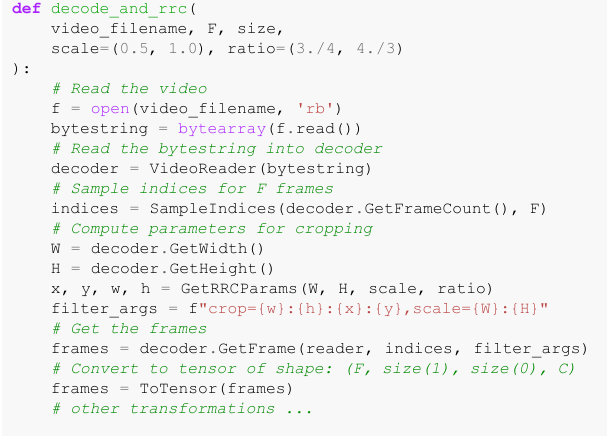}
    \caption{\small{Fused DecodeCrop.}}
	\label{fig:decode_and_rrc}
    \end{subfigure}
    \caption{\small{Pythonic pseudo-code for video decoding with a cropping filter (\shortrefsec{method:transform}).}}
    \label{fig:transform_code}
\end{figure}

\shortreffig{transform_code} illustrates the Pythonic pseudo-code for standard RandomResizedCrop for video inputs (``Decode-then-crop'') and our proposed Fused DecodeCrop.

\section{Implementation Details}
\label{sec:appdx:impl}

\subsection{Pre-training on Ego4D}
\label{sec:appdx:impl:pretrain_ego4d}
We pre-train on the video-narration pairs from Ego4D~\cite{grauman2022ego4d} with the training recipe inherited from LaViLa~\cite{zhao2022learning}.
We train the model using AdamW with $(\beta_1,\beta_2)=(0.9, 0.999)$ and weight decay of 0.01 for 5 epochs.
After Large Language Models augment the video-narrations pairs, the ``effective'' number of epochs is doubled to 10.
We use a fixed learning rate of 3e-5.
The projection head after the dual encoders is a linear layer with an output dimension of 256.

Our optimized pipeline enables us to put a per-gpu batch size of 256 on a single 8-GPU machine for ViT-B, resulting in a total batch size of 2,048.
For ViT-L, we fit a per-gpu batch size of 112 over 8 GPUs, resulting in a total batch size of 896, which is close to 1K.

For input, we randomly sample 4 frames between the start and end time of the clip and use standard \texttt{RandomResizedCrop (0.5, 1.0)}, which is fused at the video-decoding side, for data augmentation and the input resolution is $224\times 224$.

\subsection{Multi-Instance Retrieval on EK-100}
We fine-tune the pre-trained model on EK100 using AdamW with $(\beta_1,\beta_2)=(0.9, 0.999)$ and weight decay of 0.01.
We use cosine annealing with warmup, where the base learning rate starts from 1e-6, linearly increases to a peak of 3e-5 in the first epoch, and then gradually decreases to 1e-5 following a half-wave cosine schedule.
We apply the multi-instance max-margin loss~\cite{wray2019jpose} with a margin value of 0.2.
We use a per-gpu batch size of 64 over 8 GPUs for ViT-B and a per-gpu batch size of 24 over 8 GPUs for ViT-L.
We use a stochastic depth ratio of 0.1 in the backbone.

For input, we represent each video clip with 16 sampled frames at both training and testing times.
At training time, we scale the short side of the video to 256 pixels and then take a 224$\times$224 crop and use standard \texttt{RandomResizedCrop (0.5, 1.0)}, which is fused at the video-decoding side, for data augmentation.
At testing time, we scale the short side to 224 pixels and take the center 224$\times$224 crop.

\subsection{Action Recognition on EK-100}
We fine-tune the pre-trained model on EK100 for 100 epochs using SGD with a momentum of 0.9 and weight decay of 5e-4.
We use cosine annealing with warmup, where the base learning rate starts from 1e-6, linearly increases to a peak of 0.012 in the first epoch, and then gradually decreases to 1e-5 following a half-wave cosine schedule.
We drop the linear projection head and attach a $3806$-dim head for action classification.
To get the verb- and noun-level accuracy, we simply marginalize the action-level probability.

We use a per-gpu batch size of 64 over 8 GPUs for ViT-B and a per-gpu batch size of 24 over 8 GPUs for ViT-L.
We use a stochastic depth ratio of 0.1 in the backbone and apply a dropout of 0.5 before the classification head.
We also use a label smoothing of 0.1 and a mixup of 0.8.

For input, we represent each video clip with 16 sampled frames at both training and testing times.
At training time, we scale the short side of the video to 256 pixels and then take a 224$\times$224 crop and use standard \texttt{RandomResizedCrop (0.5, 1.0)} and \texttt{HorizontalFlip (0.5)}, both of which are fused at the video-decoding side, for data augmentation.
At testing time, we scale the short side to 224 pixels and take the center 224$\times$224 crop.

{
    \small
    \bibliographystyle{ieeenat_fullname}
    \bibliography{references}
}

\end{document}